\documentclass[runningheads,a4paper]{llncs}

\usepackage{amssymb}
\usepackage{graphicx}

\usepackage{graphicx}
\usepackage{epstopdf}

\usepackage{booktabs}

\usepackage{verbatim}
\usepackage{algorithm}
\usepackage{algpseudocode}
\usepackage{xspace}
\usepackage{amsmath}
\usepackage{caption}
\usepackage{subcaption}

\usepackage{rotating}
\usepackage{verbatimbox}
\usepackage[overlay,absolute]{textpos}

\begin{document}

\begin{textblock*}{10in}(38mm, 10mm)
{\textbf{Ref:} \emph{International Conference on Artificial Neural Networks (ICANN)}, Springer LNCS,}
\end{textblock*}
\begin{textblock*}{10in}(38mm, 15mm)
{Vol.~9887, pp.~170--178, Barcelona, Spain, September 2016.}
\end{textblock*}

\mainmatter  % start of an individual contribution

\title{DNN-Buddies: A Deep Neural Network-Based Estimation Metric for the Jigsaw Puzzle Problem}

\titlerunning{Deep Neural Network Based Estimation Metric for the Jigsaw Puzzle Problem}


\author{Dror Sholomon\inst{1} \and Eli (Omid) David\inst{1} \and Nathan S. Netanyahu\inst{1,2} }

\authorrunning{D. Sholomon, E.O. David, N.S. Netanyahu}

\institute{
Department of Computer Science, Bar-Ilan University, Ramat-Gan 52900, Israel \\
\email{dror.sholomon@gmail.com, mail@elidavid.com, nathan@cs.biu.ac.il}
\and
Center for Automation Research, University of Maryland, College Park, MD 20742\\
\email{nathan@cfar.umd.edu}
}

\toctitle{DNN-Buddies: Estimation Metric for the Jigsaw Puzzle Problem}
% \tocauthor{Authors' Instructions}
\maketitle

\begin{abstract}
This paper introduces the first deep neural network-based estimation metric for the jigsaw puzzle problem. Given two puzzle piece edges, the neural network predicts whether or not they should be adjacent in the correct assembly of the puzzle, using nothing but the pixels of each piece. The proposed metric exhibits an extremely high precision even though no manual feature extraction is performed. When incorporated into an existing puzzle solver, the solution's accuracy increases significantly, achieving thereby a new state-of-the-art standard.
%\keywords{We would like to encourage you to list your keywords within
%the abstract section}
\end{abstract}

\begin{figure}
\centering
         \begin{subfigure}[t]{0.45\textwidth}
                \centering
                \includegraphics[width=\textwidth]{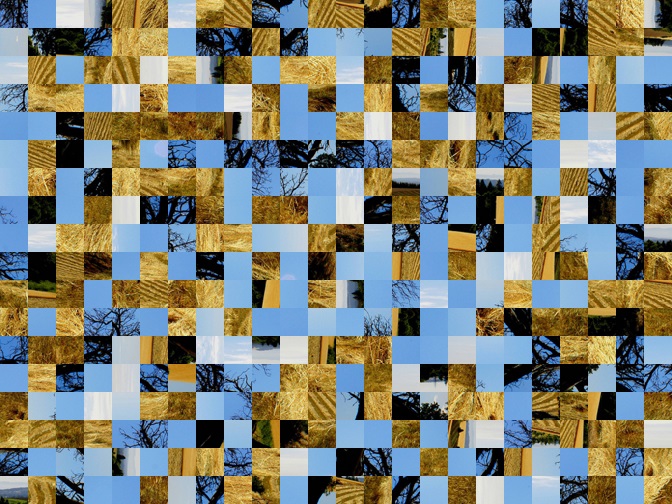}
                \caption{}
                \label{fig:intro_10375_gen_00000000}
        \end{subfigure}
        ~
        \begin{subfigure}[t]{0.45\textwidth}
                \centering
                \includegraphics[width=\textwidth]{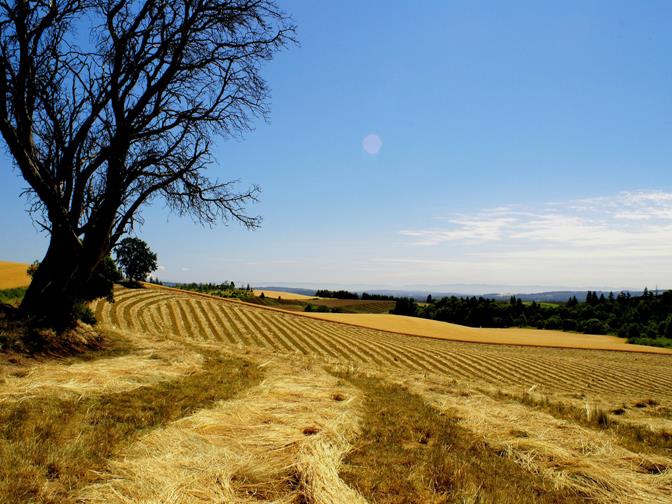}
                \caption{}
                \label{fig:intro_10375_orig}
        \end{subfigure}
        \caption{Jigsaw puzzle before and after reassembly using our DNN-Buddies scheme in an enhanced solver.}
        \label{fig:introFig}
\end{figure}

\section{Introduction}

Jigsaw puzzles are a popular form of entertainment, available in different variation of difficulty to challenge children, adults and even professional players. Given $n \times m$ different non-overlapping tiles of an image, the objective is to reconstruct the original image, taking advantage of both the shape and chromatic information of each piece. Despite the popularity and vast distribution of jigsaw puzzles, their assembly is not trivial computationally, as this problem was proven to be NP-hard~\cite{journals/aai/Altman89}~\cite{springerlink:10.1007/s00373-007-0713-4}. Nevertheless, a computational jigsaw solver may have applications in many real-world applications, such as biology~\cite{journals/science/MarandeB07}, chemistry~\cite{oai:xtcat.oclc.org:OCLCNo/ocm45147791}, literature~\cite{conf/ifip/MortonL68}, speech descrambling~\cite{Zhao:2007:PSA:1348258.1348289}, archeology~\cite{journals/tog/BrownTNBDVDRW08}~\cite{journals/KollerL06}, image editing~\cite{bb43059}, and the recovery of shredded documents or photographs~\cite{cao2010automated,conf/icip/DeeverG12,justino2006reconstructing,marques2009reconstructing}. Regardless, as noted in~\cite{GolMalBer04}, research of the topic may be justified solely due to its intriguing nature.

Recent years have witnessed a vast improvement in the research and development of automatic jigsaw puzzle solvers, manifested in both puzzle size, solution accuracy, and amount of manual human intervention required. In its most basic form, every puzzle solver requires some function to evaluate the compatibility of adjacent pieces and a strategy for placing the pieces as accurately as possible. Most strategies are greedy and rely heavily on some ``trick'' to estimate whether two pieces are truly adjacent (e.g. two pieces that are each the most compatible piece from all pieces to one another, four pieces that form a loop where each pair's compatibility is above a threshold, etc). Such heuristics were dubbed an ``estimation metric'' in~\cite{conf/cvpr/PomeranzSB11}, as they allow estimating the adjacency correctness of two pieces without knowing the correct solution. The majority of recent works focused on devising elaborate, hand-crafted compatibility functions and high-precision estimation metrics.

Despite the proven effectiveness of neural networks in the field of computer vision, no attempt has been made to automatically devise a high-precision estimation metric for the jigsaw puzzle problem. This might be due to the highly imbalanced nature of the puzzle problem, as in each $n \times m$ puzzle, there are $\mathcal{O}(n \times m)$ matching piece-pairs and $\mathcal{O}(n^2 \times m^2)$ possible mismatching ones. In this paper we propose a novel estimation metric relying on neural networks. The proposed metric achieves extremely high precision despite the lack of any manually extracted features.

The proposed metric proves to be highly effective in real-world scenarios. We incorporated the metric in our GA-based solver, using no hand-crafted sophisticated compatibility measure and experimented with the currently known challenging benchmarks of the hardest variant of the jigsaw puzzle problem: non-overlapping, $(28 \times 28)$ square pieces (i.e. only chromatic information is available to the solver) where both piece orientation and puzzle dimensions are unknown. The enhanced solver proposed sets a new state-of-the-art in terms of the accuracy of the solutions obtained and the number of perfectly reconstructed puzzles.

\section{Previous Work}

Jigsaw puzzles were first introduced around 1760 by John Spilsbury, a Londonian engraver and mapmaker. Nevertheless, the first attempt by the scientific community to computationally solve the problem is attributed to Freeman and Garder~\cite{bb47278} who in 1964 presented a solver which could handle up to nine-piece problems. Ever since then, the research focus regarding the problem has shifted from shape-based to merely color-based solvers of square-tile puzzles. In 2010 Cho \textit{et al.}~\cite{conf/cvpr/ChoAF10} presented a probabilistic puzzle solver that could handle up to 432 pieces, given some a priori knowledge of the puzzle. Their results were improved a year later by Yang \textit{et al.}~\cite{yang2011particle} who presented a particle filter-based solver. Furthermore, Pomeranz \textit{et al.}~\cite{conf/cvpr/PomeranzSB11} introduced that year, for the first time, a fully automated square jigsaw puzzle solver that could handle puzzles of up to 3,000 pieces. Gallagher~\cite{conf/cvpr/Gallagher12} has further advanced this by considering a more general variant of the problem, where neither piece orientation nor puzzle dimensions are known. Son \textit{et al.}~\cite{son2014solving} improved the accuracy of the latter variant using so-called ``loop-constraints''. Palkin and Tal~\cite{paikin2015solving} further improved the accuracy and handled puzzles with missing pieces. Sholomon \textit{et al.}~\cite{Sholomon_2013_CVPR} presented a genetic algorithm (GA)-based solver for puzzles of known orientation which was later generalized to other variants~\cite{sholomon2014generalized,sholomon2014genetic}.

\subsection{Compatibility Measures and Estimation Metrics}

As stated earlier, most works focus on the compatibility measure and an estimation metric. A compatibility measure is a function that given two puzzle piece edges (e.g. the right edge of piece 7 versus the upper edge of piece 12) predicts the likelihood that these two edges are indeed placed as neighbors in the correct solution. This measure applies to each possible pair of piece edges. The estimation metric, on the other hand, predict whether two piece edges are adjacent but may not apply to many possible pairs. Following is a more detailed review of the efforts made so far in the field.

Cho \textit{et al.}~\cite{conf/cvpr/ChoAF10} surveyed four compatibility measures among which they found dissimilarity the most accurate.  Dissimilarity is the sum (over all neighboring pixels) of squared color differences (over all color bands). Assuming pieces $x_{i}$, $x_{j}$ are represented in some three-dimensional color space (like RGB or YUV) by a $K \times K \times 3$ matrix, where $K$ is the height/width of a piece (in pixels), their dissimilarity, where $x_{j}$ is to the right of $x_{i}$, for example, is
\begin{equation} \label{eq:dissimilarity}
D(x_{i},x_{j},r)=\sqrt{\sum_{k=1}^{K}\sum_{cb=1}^{3}(x_{i}(k,K,cb)-x_{j}(k,1,cb))^{2}},
\end{equation}
where $cb$ denotes the color band.

Pomeranz \textit{et al.}~\cite{conf/cvpr/PomeranzSB11} also used the dissimilarity measure but found empirically that using the $(L_p)^q$ norm works better than the usual $L_2$ norm. Moreover, they presented the high-precision {\em best-buddy} metric. Pieces $x_{i}$ and $x_{j}$ are said to best-buddies if
\begin{align}
\forall x_{k} \in Pieces, \; C(x_{i},x_{j},R_1) \geq C(x_{i},x_{k},R_1)\notag \\
\text{and \quad\quad\quad\quad\quad\quad\quad\quad}\\
\forall x_{p} \in Pieces, \; C(x_{j},x_{i},R_2) \geq C(x_{j},x_{p},R_2) \notag
\end{align}
where $Pieces$ is the set of all given image pieces and $R_1$ and $R_2$ are ``complementary'' spatial relations (e.g. if $R_1$ = right, then $R_2$ = left and vice versa).
 
Gallagher~\cite{conf/cvpr/Gallagher12} proposed yet another compatibility measure, called the {\em Mahalanobis gradient compatibility} (MGC) as a preferable compatibility measure to those used by Pomeranz \textit{et al.}~\cite{conf/cvpr/PomeranzSB11}. The MGC penalizes changes in intensity gradients, rather than changes in intensity, and learns the covariance of the color channels, using the Mahalanobis distance. Also, Gallagher suggested using {\em dissimilarity ratios}. Absolute distances between potential piece edge matches are sometimes not indicative (for example in smooth surfaces like sea and sky), so considering the absolute score, divided by the second-best score available seems more indicative.

Son \textit{et al.}~\cite{son2014solving} suggested ``loop-constraints'', four or more puzzle piece edges where the compatibility ratio between each pair is in the top ten  among all possible pairs of piece edges in the given puzzle. Palkin and Tal~\cite{paikin2015solving} proposed a greedy solver based on an $L_1$-norm asymmetric dissimilarity and the best-buddies estimation metric.

\section{DNN-Buddies}

\subsection{Motivation}

We propose a novel estimation metric called ``DNN-Buddies''. Our goal is to obtain a classifier which predicts the adjacency likelihood of two puzzle piece edges in the correct puzzle configuration.

Note that despite the exponential nature of the problem (as there are $\mathcal{O}((nm)!)$ possible arrangements of the pieces, taking into account rotations), the problem can be solved theoretically by assigning correctly, in a consecutive manner, $n \times m - 1$ piece-edge pairs. (This is reminiscent of finding a minimal spanning tree, as noted by~\cite{conf/cvpr/Gallagher12}.) Hence, the classifier's precision is of far greater importance than its recall. A classifier with perfect precision and a recall of
\begin{equation}
\frac{n \times m - 1}{\text{all possible matches}} = \frac{n \times m - 1}{4 \times (n \times (m-1) + (n-1) \times m)} < \frac{1}{8}
\end{equation}
might achieve a perfect solution by itself.

\subsection{Challenges}

A straight-forward solution might have been to train a neural network against matching-pairs vs. non-matching ones. However, the issue of a jigsaw puzzle piece matching is of an imbalanced nature. In each $n \times m$ puzzle, there are $\mathcal{O}(n \times m)$ matching pairs of piece edges and $\mathcal{O}(n^2 \times m^2)$ possible nonmatching ones. A thorough review on the challenges and tactics to avoid  them can be found in~\cite{he2009learning}.

The trivial approach of random or uninformed undersampling, i.e. randomly choosing the required number of nonmatching pairs leads to a low-precision and high-recall metric, the very opposite of the goal set beforehand. We believe that the reason for this shortcoming is that there exist many ``easy-to-spot'' mismatches but only a handful of ``hard-to-spot'' ones. Thus, we resort to informed undersampling, choosing a subset of ``good'' mismatching pairs according to some criterion. Nevertheless, we avoid using any manual feature selection or other sophisticated image-related means.

In the jigsaw puzzle domain, similarly to many other problem domains, the solver does not actually try to reassemble the original image (as this problem is not mathematically defined), but rather tries solving a ``proxy problem'' which is to achieve an image whose global overall score between abutting-edges is minimal. Thus, we choose using the compatibility measure as the undersampling criterion.

\subsection{Neural Network Training}

For training and cross-validation, we use the 2,755 images of size $360 \times 480$ pixels from the IAPR TC-12 Benchmark~\cite{grubinger2006iapr}. Each image is first converted to YUV space followed by the normalization of each channel separately (via $z$-score normalization). Next, each (puzzle) image is divided to $12 \times 17$ tiles, where each tile is of size $28 \times 28$ pixels (as in all previous works); finally, we create a balanced set of positive and negative samples of puzzle-piece pairs, using informed undersampling as will be described below. In the end, we obtain a balanced set of 970,224 pairs overall. 

To balance our dataset, we use the most basic compatibility score which is the dissimilarity between two piece-edges in the YUV color-space, as described in Eq.~\ref{eq:dissimilarity}, as an undersampling criterion. For each puzzle piece edge $x_{i,j} (i = 1..n \times m, j = 1..4)$, we find its most compatible piece edge $x_{k1,l1}$ and its second most compatible piece edge $x_{k2,l2}$. If the pair of edges $x_{i,j}-x_{k1,l1}$ is indeed adjacent in the original image, we add this pair to the pool of positively-labeled samples and toss the pair $x_{i,j}-x_{k2,l2}$ to the pool of negatively-labeled samples. Otherwise, $x_{i,j}-x_{k1,l1}$ is added to the negatively-labeled samples and the other pair is discarded. The latter is done to avoid training the network on adjacent pieces which happen to be vastly different due to a significant change of the image scenery in the corresponding region. In other words, we restrict our interest to highly compatible piece edges that are indeed adjacent. Since this method leads to more negative samples than positive ones, we eventually randomly throw some negative samples to balance out the set.

\begin{figure}[t]
	\centering
	\includegraphics[height=1.5in]{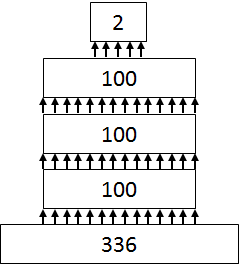}
	\caption{Architecture of our DNN-Buddies scheme.}
	\label{fig:ffnn}
\end{figure}

From each image pair we extract the two columns near the edge, i.e. the column of abutting pixels in each edge and the one next to it. This results is an input of size $(28 \times 4 \times 3 = ) ~336$ pixels. We use a feed-forward neural network (FFNN) of five fully connected layers of size 336, 100, 100, 100, and 2. The output is a softmax layer containing two neurons. We expect (0,~1) for matching pairs and (1,~0) otherwise. The activation used in all layers is the rectified linear unit (ReLU) function, i.e.  $f(x) = max(0, x)$. Figure~\ref{fig:ffnn} depicts the network's structure.

We trained the network in a supervised manner using {\em Stochastic Gradient Descent} that minimizes the negative log likelihood of the error for 100 iterations. The resulting network reaches 95.04\% accuracy on the training set and 94.62\% on a held-out test set.

All dataset preparation and network training was performed using Torch7~\cite{collobert2011torch7}.

\section {Experimental Results}

For each piece edge $x_{i,j} (i = 1..n \times m, j = 1..4)$, if its most compatible piece edge $x_{k,l}$ is classified positively using the DNN-Buddies network, we define $x_{k,l}$ to be $x_{i,j}$'s {\em DNN-buddy} piece edge. Note that each piece edge can have only a single DNN-buddy; also, some pieces might not have a DNN-buddy at all (if the most compatible piece is not classified as one by the DNN-Buddies network).

First, we evaluate the precision of the proposed metric, i.e. how many DNN-buddies are indeed adjacent in the original image. Using the well known dataset presented by Cho \textit{et al.}~\cite{conf/cvpr/ChoAF10} of 20 432-piece puzzles, we obtained a precision of 94.83\%.

Next, we incorporated the estimation metric (due to the proposed DNN-Buddies scheme) into the GA-based solver proposed by us previously ~\cite{sholomon2014genetic}. Unfortunately, due to lack of space, no self-contained review of genetic algorithms and the proposed method can be included in this paper. Nevertheless, the modification required with respect to the existing GA framework is rather simple; if a DNN-buddy pair appears in one of the parents, assign this pair in the child. Figure~\ref{fig:crossoverOverview} describes the modified crossover operator in the GA framework according to the above (see Step 2, which includes the new DNN-buddy phase).

\begin{figure}[htbp]
\begin{center}
\hrule
\bigskip
\begin{description}

\item \textbf{Until} $(n - 1)$ relative relations are assigned \textbf{do}

\begin{enumerate}

\item Try assigning all {\em common} relative relations in the parents.
\item Try assigning all {\em DNN-buddy} relative relations in the parents.
\item Try assigning all {\em best-buddy} relative relations in the parents.
\item Try assigning all existing {\em most-compatible} relative relations.
\item Try assigning {\em random} relative relations.
\end{enumerate}

\end{description}
\bigskip
\hrule
\bigskip

\caption{Crossover overview}
\label{fig:crossoverOverview}
\end{center}
\end{figure}

We ran the augmented solver on the 432-piece puzzle set and on the two additional datasets proposed by Pomeranz \textit{et al.}~\cite{conf/cvpr/PomeranzSB11} of 540- and 805- piece puzzles. We evaluated our results according to the {\em neighbor comparison} which measures the fraction of correct neighbors and the number of puzzles perfectly reconstructed for each set.

Table~\ref{tab:results} presents the accuracy results of the same solver with and without the DNN-Buddies metric. For each dataset we achieve a considerable improvement in the overall accuracy of the solution, as well as the number of perfectly reconstructed puzzles. Moreover, our enhanced deep neural network-based scheme appears to outperform the current state-of-the-art results, as it yields accuracy levels of 95.65\%, 96.37\% and 95.86\%, which surpass, respectively, the best results known of 95.4\%~\cite{paikin2015solving}, 94.08\% and 94.12\%~\cite{sholomon2014genetic}.

\begin{table}
\centering
\setlength{\tabcolsep}{1em} % for the horizontal padding
{\renewcommand{\arraystretch}{1.1}% for the vertical padding
\begin{tabular}{ |c||c|c||c|c| }
    \hline
    & \multicolumn{2}{|c||}{GA} & \multicolumn{2}{|c|}{Our (GA + DNN-Buddies)} \\ \hline
  \# of Pieces  & Neighbor & Perfect & Neighbor & perfect \\ \hline \hline
  432 & 94.88\% & 11 & 95.65\% & 12\\ \hline
  540 & 94.08\% & 8  & 96.37\% & 11 \\ \hline
  805 & 94.12\% & 6 & 95.86\% & 8 \\ \hline

\end{tabular}
}
\medskip
\caption{Comparison of our accuracy results with and without the new DNN-Buddies estimation metric.}
\label{tab:results}
\end{table}

\vspace*{-20pt} 
\section{Conclusions}

In this paper we presented the first neural network-based estimation metric for the jigsaw puzzle problem. Unlike previous methods, no manual feature crafting was employed. The novel method exhibits high precision and when combined with a real-world puzzle solver, it significantly improves the solution's accuracy to set a new state-of-the art standard.

\bibliographystyle{splncs03}

\end{document}